\documentclass[affilfootnotes=true]{agn_article}
\usepackage{hyperref}
\usepackage[backend=biber,style=numeric]{biblatex}
\usepackage{amsmath}
\usepackage{amssymb}
\usepackage{graphicx}
\usepackage{wrapfig}
\usepackage{float}
\DeclareUnicodeCharacter{0301}{\'{e}}

\defineaffiliation{agn}{%
	Mathematics for Engineers,
	Institue for Technologies of Metals,
	University of Duisburg-Essen,
	Friedrich-Ebert-Str.~12,
	47119 Duisburg, Germany%
}%

\title{Interpretation of multi-label classification models using shapley values}
\author{%
	Chen Shikun\affil{agn}\corresponding{shikun.chen@stud.uni-due.de}%
}

\date{\today}
\begin{document}
\maketitle
\begin{abstract}
Multi-label classification is a type of classification task, it is used when there are two or more classes, and the data point we want to predict may belong to none of the classes or all of them at the same time. In the real world, many applications are actually multi-label involved, including information retrieval, multimedia content annotation, web mining, and so on. A game theory-based framework known as SHapley Additive exPlanations (SHAP) has been applied to explain various supervised learning models without being aware of the exact model. Herein, this work further extends the explanation of multi-label classification task by using the SHAP methodology. The experiment demonstrates a comprehensive comparision of different algorithms on well known multi-label datasets and shows the usefulness of the interpretation.
\end{abstract}
\smallskip
\smallskip
\section{Introduction}
A multi-label classification deals with problems with a situation where a data point can belong to more than one class. \cite{charte2016multilabel} The form of output is the essential difference between tradition and multi-label classification. In contrast to tradition classification problem, in multi-label learning each object is also represented by a single instance while associated with a set of labels instead of a single label. There are many scenarios that can be resovled by employing multi-label classification. For example, news can be categorized into multiple topics, music or film can be assigned with more than one genre. \par
The rapidly increasing usage of Machine Learning (ML) raises the attention of explainable ML to help people better trust and understand models at scale. As humans, we must be able to fully understand how decisions are being made so that we can trust the decision of ML systems. In principle, we need ML models to function as expected, to produce transparent explanations, and to be visible in how they work. \par
Recently, many methods have been developed to interpret predictions from supervised ML models. LIME \cite{ribeiro2016should} is an approach can explain the predictions of any classifiers by learning an interpretable model locally around the prediction. DeepLIFT \cite{shrikumar2017learning} is an example of a model-specific method for explaining deep
learning models in which the contributions of all neurons in the network are back-propagated to the input features \cite{antwarg2020explaining}. However, the SHAP \cite{lundberg2017unified} inspired by previous methods and other several methods \cite{vstrumbelj2014explaining,datta2016algorithmic,bach2015pixel,lipovetsky2001analysis} was proposed as a united approach to explain the output of any ML model. \par
This paper introduces the interpretability of multi-label classification problem by conducting three algorithms on three real-world datasets. The different aspects of SHAP framework were also compared. The remainder of the paper is structured as follows. First, some background knowledge about multi-label classification and SHAP are briefy described in 
\hyperref[sec:background]{Section 2}. Next, \hyperref[sec:experiment]{Section 3} presents the setup of the experiments to show the interpretability of multi-label classification problem. The results of these experiments are then showed and discussed in \hyperref[sec:results]{Section 4}. Finally, some directions for future work are discussed and the paper is concluded in \hyperref[sec:summary]{Section 5}.

\section{Background}
\label{sec:background}
There are two main approaches to solve a multi-label classification problem: \textit{algorithm adaptation} and \textit{problem transformation} \cite{zhang2013review}.
This section describes three current state-of-the art learning algorithms to multi-label classification: BinaryRelevance (BR) and ClassifierChain (CC) which are problem transformation algorithms, and MLKNN is the algorithm adaptation category. The principles of the SHAP methodology and its advantages will also be covered. \par
\subsection{BinaryRelevance}
Binary relevance is the most trivial approach to multi-label classification \cite{boutell2004learning}. It transforms a multi-label classification problem with \textit{L} labels into \textit{L} single-label separate binary classification problems using the same base classifier provided. Binary relevance takes adavantage of adopting a \textit{one-vs-all} ensemble method, training independent binary classifiers to predict the relevance of each label to a data sample. The independent predictions are then aggregated to form a set of relevant lables \cite{pakrashi2016benchmarking}. 
\par
\subsection{ClassifierChain}
Classifier Chain \cite{read2011classifier} treats each label as a part of conditional chain of single-class classification problems. It is a \textit{one-vs-all} classifier built for each label, however, these classifiers are chained together so that the outputs of classifiers early in the chain are used as inputs into sebsequent classifiers according to \textit{Bayesian chain rule}.
\subsection{MLKNN}
Multi-label K-nearest neighbor (MLKNN) \cite{zhang2007ml} was the first lazy approach developed sepcifically for multi-label classification, and it is one of the most widely cited algorithm adaptation approaches. The basic idea of MLKNN algorithm is to adapt \textit{k-nearest neighbor} techniques to deal with multi-label data, where  \textit{Maximum A Posteriori}(MAP) principle is utilized to make prediction by reasoning with the labeling information embodied in the neighbors \cite{zhang2007ml}. 
\subsection{SHAP}
The Shapley value (SHAP) was originally proposed to estimate the importance of an individual player in a collaborative team \cite{shapley1953value, lundberg2017unified}. SHAP interprets the Shapley value as an additive feature attribtuion method, and SHAP interprets the predicted value of the model as the sum of the attribution value of each input feature:
\begin{equation}
g(z') = \phi_{0} + \sum^{M}_{j=1}\phi_{j}z'_{j} 
\end{equation}
where $g$ is interpretation model, $z'\in \{0,1\}^M$ indicates whether the corresponding feature can be observed (1 or 0). $M$ is the number of input, and $\phi_{i}\in \mathbb{R}$ is the attribution value (Shapley value) of each feature. $\phi_{0}$ is the constant of interpretation model (namely the predicted mean value of all training samples)\cite{lundberg2017unified}. \par
Lundberg and Lee \cite{lundberg2017unified} demonstrate that SHAP is better aligned with human intuition than other ML interpretation methods. There are three benefits worth mentioning here.
\begin{enumerate}
\item The first one is \textit{global interpretability} -- the collective SHAP values can show how much each predictor contributes, either positively or negatively, to the target variable. The features with positive sign contribute to the final prediction activity, whereas features with negative sign contribute to the prediction inactivity. In particular, the importance of a feature $i$ is defined by the SHAP as follow:
\begin{equation}\label{eq:2}
\phi_{i} = \frac{1}{|N|!}\sum_{S\subseteq {N} \setminus \{i\}} |S|!(|N| - |S| -1)![f(S \cup \{i\}) - f(S) ]
\end{equation}
Here $f(S)$ is the output of the ML model to be interpreted using a set of $S$ of features, and $N$ is the complete set of all features. The contribution of feature $i(\phi_{i})$ is determined as the average of its contribution among all possible permutations of a feature set. Furthermore, this equation considers the order of feature, which influence the observed changes in a model's output in the presence of correlated features \cite{rodriguez2020interpretation}. 
\item The second benefit is \textit{local interpretability} -- each observation gets its own set of SHAP values. This greatly increase its transparency. We can explain why a case receives its prediction and the contributions of the predictors. Traditional variable importance algorithms only show the results across the entire population but not on each individual case. The local interpretability enables us to pinpoint and contrast the impact of the factors.
\item Third, SHAP framework suggests a model-agnostic approximation for SHAP values, which can be calculated for any ML model, while other methods use linear regression or logistic regression models as the surrogate models. 

\end{enumerate}

\section{Experiment}
\label{sec:experiment}
To acess the interpretability of multi-label classification, experiments were performed using 3 well known multi-label datasets, listed in \hyperref[tab:table1] {Table 1}. \textit{Instances, Inputs} and \textit{Labels} indicate the total number of data points, the number of predictor variables, and the number of potential labels, in each dataset respectively. \par

\begin{table}[]
\begin{center}
\caption {The datasets used in the experiments described in this paper and their properties} \label{tab:table1} 
\smallskip
\begin{tabular}{lllllllll}
\multicolumn{1}{c}{Dataset} & \multicolumn{1}{c}{Domain}   & \multicolumn{1}{c}{Instances} & \multicolumn{1}{c}{Inputs} & \multicolumn{1}{c}{Labels} \\ \hline
\multicolumn{1}{c}{yeast} & \multicolumn{1}{c}{biology} & \multicolumn{1}{c}{2417}     & \multicolumn{1}{c}{103}   & \multicolumn{1}{c}{14}   \\ \hline
\multicolumn{1}{c}{water quality} & \multicolumn{1}{c}{chemical} & \multicolumn{1}{c}{1060}     & \multicolumn{1}{c}{16}   & \multicolumn{1}{c}{14}   \\ \hline
\multicolumn{1}{c}{foodtruck} & \multicolumn{1}{c}{information} & \multicolumn{1}{c}{407}     & \multicolumn{1}{c}{21}   & \multicolumn{1}{c}{12}   \\ \hline
\end{tabular}
\end{center}
\end{table}
Classifiers BR, CC and MLKNN are all used to deal with these 3 multi-label datasets. Although the performance of the algorithms is not key objective to this study. A wide range of hyper-parameters were explored to attain the relatively best performance. For each hyper-parameter combination,a 2 \text{x} 5-fold cross-validation of grid search was performed for each dataset. The final hyper-parameters used for each algorithm and its corresponding dataset are described as follows. For BR classifier, a \textit{RandomForestClassifier} with $criterion=entropy, max\_depth=15, min\_samples\_leaf=2$ was chosen as a base classifier to train the all three datasets. For CC classifier,  a \textit{RandomForestClassifer} with $criterion=entropy, max\_depth=3, order=random$ was selected as base classifier to train three datasets. For MLKNN, the hyper-parameter $k$ determines the number of neighbors of each input instance to take into account, a total of 20 values of $k$ were explored, where $k \in \{1,2,3,\dots,20\}$. To train the all three datasets, $k=5$ was chosen as optimal hyper-parameter. \par
This work focus on exploiting the interpretability of multi-label classification, especially to local and global interpretation. Therefore does not train a model to achieve benchmark results. To explain the output of ML model, SHAP provides two basic functions, kernel and tree. Kernel SHAP approximates Eq. \hyperref[eq:2]{2}, it is subject to sampling variability, and requires a background data set for training. Although tree SHAP is much faster than kernel SHAP, however, it can be only used for tree-based models. In this work, the kernel SHAP is applied to explain ML predictions. 
\section{Results}
\label{sec:results}
Global and local are two scopes of interpretation. Global model interpretation means understand the distribution of the prediction output based on the features, answering the question how does the trained model make predictions. The SHAP framework provides two ways to visulize global interpretation, feature importance plot and summary plot. On the other hand, local interpretation focus specifically on a single prediction or a group of predictions, and try to understand model decisions for that point based on local region. Local data distributions and feature spaces might behave completely different and give more accurate explanations on opposed to global interpretations \cite{molnar2020interpretable}. We can use force plot to show individual SHAP value for some specific observation. For all three datasets and their visulization results on three algorithms will be shown next. 

\subsection{Feature Importance}\label{subsec:featureimportance}
\begin{figure}[h]
\centering
\includegraphics[width=\textwidth]{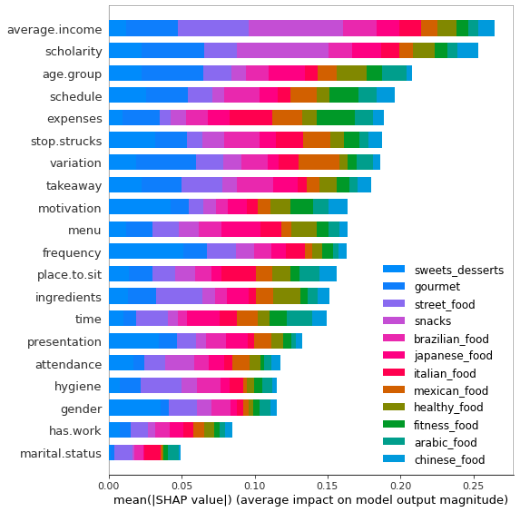}
\caption{SHAP feature importance for the foodtruck dataset trained with BR}
\label{fig:1}
\end{figure}

The idea behind SHAP feature importance is simple: Features with large absolute Shapley values are important. After calculating the absolute Shapley values per feature across the data, we sort the features by decreasing importance. To demonstrate the SHAP feature importance, we take foodtruck as the example. \hyperref[fig:1] {Figure 1} shows the SHAP feature importance for the foodtruck dataset which trained with BR.\par

\begin{figure}[h]
\centering
\includegraphics[width=\textwidth]{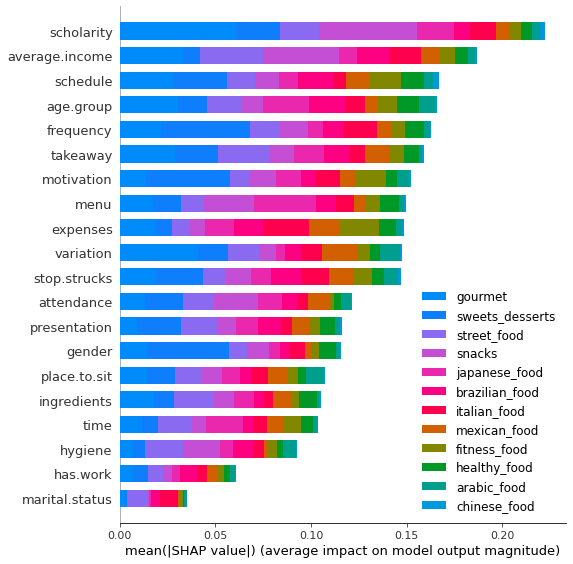}
\caption{SHAP feature importance for the foodtruck dataset trained with CC}
\label{fig:2}
\end{figure}

\begin{figure}[h]
\centering
\includegraphics[width=\textwidth]{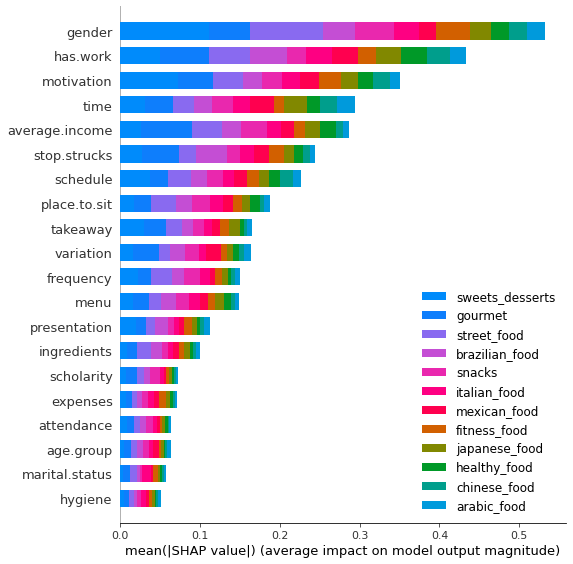}
\caption{SHAP feature importance for the foodtruck dataset trained with MLKNN}
\label{fig:3}
\end{figure}

The dataset contains 12 labels to predict. It indicates that \texttt{averageincome} is the most important feature followed by attribute \texttt{scholarity} and \texttt{agegroup}. Furthermore, the impact of features on individual label is also different. For example, the attribute \texttt{averageincome} has large impact on label \texttt{snacks} but small impact on label \texttt{chinese\_food}. Similarly, the SHAP feature importance for foodtruck dataset trained with CC and MLKNN are presented \hyperref[fig:2]{Figure 2} and \hyperref[fig:3]{Figure 3}. It is worth noting that the top 4 most importance features of BR and CC are almost identical. However, the feature importance of MLKNN shows a totally different ranking. 

\subsection{Summary Plot}\label{subsec:summaryplot}

The summary plot combines feature importance with feature effects. Each point on the summary plot is a Shapley value for a feature and an instance. The position on the y-axis is determined by the feature and on the x-axis by the Shapley value. The color represents the value of the feature from low to high. Overlapping points are jittered in y-axis direction, so we get a sense of the distribution of the Shapley values per feature. The features are ordered according to their importance \cite{molnar2020interpretable}.

\begin{figure}[H]
\centering
\includegraphics[width=\textwidth]{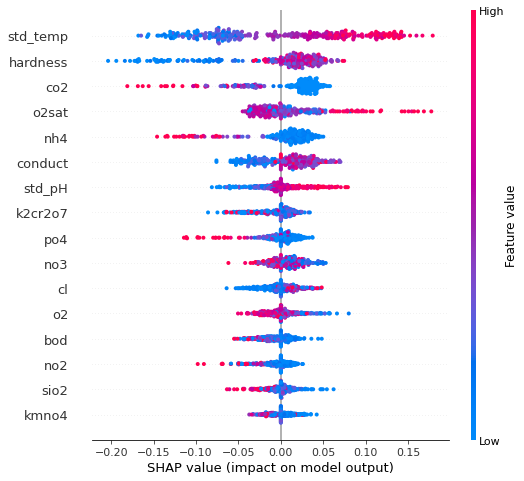}
\caption{SHAP summary plot for the water quality dataset trained with BR}
\label{fig:4}
\end{figure}
\begin{figure}[h]
\centering
\includegraphics[width=\textwidth]{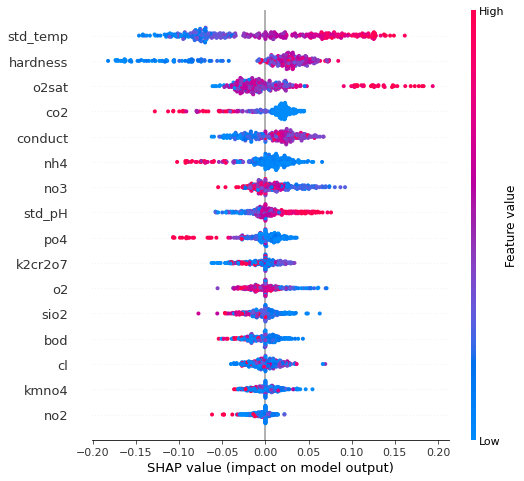}
\caption{SHAP summary plot for the water quality dataset trained with CC}
\label{fig:5}
\end{figure}
\begin{figure}[h]
\centering
\includegraphics[width=\textwidth]{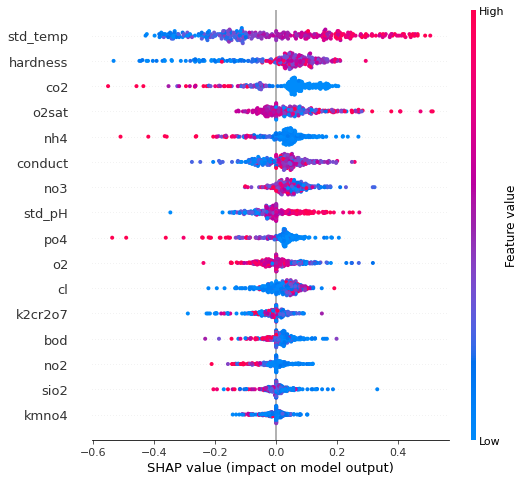}
\caption{SHAP summary plot for the water quality dataset trained with MLKNN}
\label{fig:6}
\end{figure}

To show the result of SHAP summary plot, we use the dataset water quality as example, it contains 14 labels to predict. Therefore, we only shows the summary plot on the first label. From \hyperref[fig:4] {Figure 4} we can tell that a high level of the attribute \texttt{std\_temp} content has a high and positive impact on the water quality. The high comes from red color, and positive is shown on the x-axis. Similarly, we will say the attribute \texttt{co2} is negatively correlated with the target variable. It is very aligned with the intuition of human beings. The summary plot for CC and MLKNN are shown \hyperref[fig:5]{Figure 5} and \hyperref[fig:6]{Figure 6}. The indications of the relationship between the value of a feature and the impact on the prediction for these three algorithms are very similar to each other.

\subsection{Local Interpretation}\label{subsec:localinterpretation}

\begin{figure}[H]
\centering
\includegraphics[width=\textwidth]{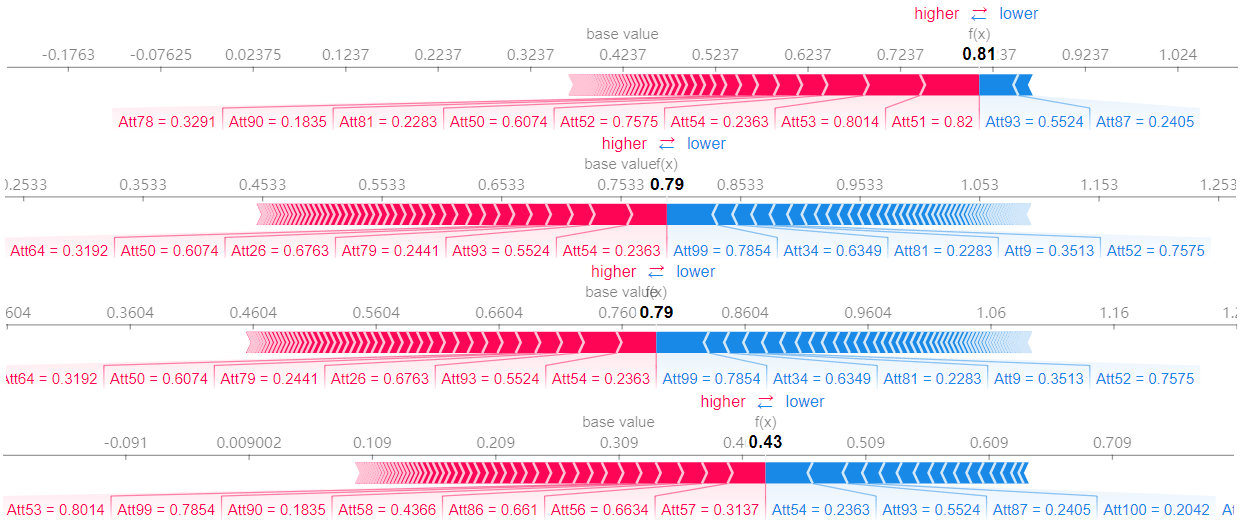}
\caption{Local interpretation for yeast dataset of labels Class 2, Class 13, Class 12, Class 1 of sample 550}
\label{fig:7}
\end{figure}

SHAP local interpretation visualizes feature attributions such as Shapley values as \texttt{forces}. Each feature value is a force that either increase or decreases the prediction. The prediction starts from the baseline. The baseline for Shapley values is the average of all predictions \cite{molnar2020interpretable}. In the \hyperref[fig:7]{Figure 7} the dataset yeast is used to demonstrate the visualization result of local interpretation. Because of large instances and many labels involved, sample 550 was randomly chosen to show its impact on lables (1, 2, 12, 13). \par

In the plot, each Shapley value is an arrow that pushes to increase (positive value) or decrease (negative value) to prediction. These forces balance each other out at the actual predition of the data instance. For instance, for the class 13 (the second one), the baseline is 0.7533. The sample 550 has a prediction of 0.79. The \texttt{Att54}, \texttt{Att26}, \texttt{Att79}, \texttt{Att93},  \texttt{Att50} can increase the prediction. On the other hand, the \texttt{Att99}, \texttt{Att34}, \texttt{Att81}, \texttt{Att9}, \texttt{Att52} can decrease the final output prediction.   

\section{Summary and Conclusion}
\label{sec:summary}
In this paper the possbility of interpretability of multi-label classification problem was proposed. The SHAP value enables the interpretation of ML models and their predictions, yielding feature importance values for individual prediction from any ML model. The necessity of model interpretability can sometimes more desiable than accuracy for real-world ML applications. The results motivate the author to extend the method towards another interpretation method (like LIME) on multi-label classification problem. \par
Although the interpretability of multi-label classification is possible, there are still many works need to be done in future. First, the relevance between features was omitted by ML algorithms, which means we may lose the correlation between features. Second, the kernel SHAP method is very slow when training a large dataset. Third, the interpreters of this work helped us open the black box, but the comparison and evaluation of existing interpretation methods still require more research. Finally, choosing a more friendly interpretation model based on the type of audience of model decision problems and standarzing the evaluation of interpretation given by the interpreter still needs more attention. 
\printbibliography
\end{document}